\documentclass[runningheads]{llncs}

\usepackage[T1]{fontenc} 
\usepackage{graphicx} 
\usepackage{booktabs} 
\usepackage{array} 
\usepackage{hyperref} 
\usepackage{doi} 
\usepackage[figuresleft]{rotating}
\usepackage[english]{babel}
\usepackage{cite}
\usepackage{orcidlink}

\usepackage[final]{microtype}
\usepackage{csquotes} 
\begin{document}
%
%
\titlerunning{Building Autonomous LLM Agents}
%

\title{
Fundamentals of Building Autonomous LLM Agents
\thanks{This paper is based on a seminar technical report from the course~\textit{Trends in Autonomous Agents: Advances in Architecture and Practice} offered at TUM.}
}
\author{
Victor de Lamo Castrillo\inst{1}\orcidlink{0009-0000-5318-1464},
Habtom Kahsay Gidey\inst{2}\orcidlink{0000-0001-5802-2606} \and
Alexander Lenz\inst{2}\orcidlink{?} \and
Alois Knoll\inst{2}\orcidlink{0000-0003-4840-076X}
}

\authorrunning{de Lamo et al.}
%
\institute{
Universitat Politècnica de Catalunya, Barcelona, Spain\\
\email{victor.de.lamo@estudiantat.upc.edu} \and
Technische Universität München, München, Germany \\
\email{\{habtom.gidey, alex.lenz, knoll\}@tum.de}
}

%
%

\maketitle
%
%
%
%
\begin{abstract}
This paper reviews the architecture and implementation methods of agents powered by large language models (LLMs). 
Motivated by the limitations of traditional LLMs in real-world tasks, the research aims to explore patterns to develop \enquote{agentic} LLMs that can automate complex tasks and bridge the performance gap with human capabilities. 
Key components include a perception system that converts environmental percepts into meaningful representations; a reasoning system that formulates plans, adapts to feedback, and evaluates actions through different techniques like Chain-of-Thought and Tree-of-Thought; a memory system that retains knowledge through both short-term and long-term mechanisms; and an execution system that translates internal decisions into concrete actions. 
This paper shows how integrating these systems leads to more capable and generalized software bots that mimic human cognitive processes for autonomous and intelligent behavior.

\end{abstract}

\section{Introduction}
\subsection{Motivation}
Artificial intelligence (AI) is a powerful technology that is transforming cognitive automation and fundamentally reshaping the way tasks are performed~\cite{gidey2023user,macedo2024evolving,gidey2023towards}. Today, one can develop remarkable systems without the need to write complex algorithms or master low-level code. We are closer than ever to realizing the idea that~\enquote{if you can think it, you can build it.} 
Instead of relying solely on programming skills, what increasingly matters is understanding how a human would reason through a problem, since LLM agents can learn and mimic human problem solving by externalizing intermediate reasoning and refining it through self-feedback~\cite{wei2023chainofthought,yao2023tree,wang2023selfconsistency,yao2023react,shinn2023reflexion,madaan2023selfrefine,huang2024understanding}.

LLM agents represent a new paradigm that breaks traditional barriers. They enable the execution of tasks that were previously costly, time-consuming, or even infeasible. More than tools, agents act as collaborators, assisting humans in dynamic environments and automating decision-making in critical systems. However, this transformation is still in its early stages. Engaging with LLM agents is comparable to engaging with a new species, one that we are only beginning to understand, train, and guide~\cite{AnthropicAgents}.

This raises a crucial question: How can we build agents who think and act intelligently? How should we structure their~\enquote*{minds} so that they can interpret information, reason, plan effectively, and make decisions that we can trust?
Building on this vision of LLM agents as intelligent collaborators, this review explores and defines the architectural foundations that enable their autonomous and effective performance in complex tasks~\cite{gidey2017grounded}.

\subsection{Review Objective}
The primary objective of this research is to review the design and implementation of intelligent agents powered by large language models (LLMs) to improve the execution of complex automation tasks~\cite{gidey2023user,gidey2023towards}. Specifically, the review focuses on the agents' perception, memory, reasoning, planning, and execution capabilities. The review aims to accomplish this by pursuing the following particular goals:

\begin{enumerate}
\setlength{\itemsep}{0.3em}
    \item Explore the options for perception systems, including multimodal LLMs and image processing tools, analyzing their contributions to interpreting visual inputs for task execution.
    \item Examine reasoning architectures, such as Chain-of-Thought (CoT) and Tree-of-Thought (ToT), and their contributions to generating structured plans for complex tasks, including how reflection enhances iterative problem solving.
    \item Explore and evaluate memory-augmented architectures, such as Retrieval-Augmented Generation (RAG) and long-term memory systems, investigating effective methods for information storage to enable practical and useful applications.
    \item Examine the available execution architectures, such as tool-based frameworks, and code generation approaches, exploring their contributions to automating tasks.
    \item Finally, evaluate the complexity of implementation of each system solution proposed.
\end{enumerate}

To achieve these objectives, some challenges need to be overcome.

\subsection{Problem Statement}

Building LLM agents to automate complex tasks can offer useful opportunities but also pose complex challenges~\cite{gidey2023towards,han2025llm,xi2023risepotential}. 
Despite all the advances in LLMs, developing agents that perform well in various scenarios remains a significant challenge~\cite{han2025llm}. 
The purpose of this study is to address these issues by reviewing each system's implementation options, assessing their contributions, and contrasting various strategies.

Benchmarks such as OSworld~\cite{osworld2024}, alongside studies on autonomous software agents~\cite{gidey2023towards, gidey2025affordance, gidey2025visual}, reveal key limitations in multimodal agents, highlighting the following issues:

\begin{enumerate}
\setlength{\itemsep}{0.3em}
    \item \textbf{Difficulties in GUI grounding and operational knowledge:} Agents struggle to accurately map screenshots to precise coordinates for their actions and lack deep understanding of basic graphical user interface (GUI) interactions and application-specific features.
    \item \textbf{Repetitive actions:} Agents frequently predict repetitive actions, indicating a lack of progress or an inability to break out of loops.
    \item \textbf{Inability to handle unexpected window noise:} Agents are not robust to unexpected elements or changes in UI layout, such as unanticipated pop-up windows or dialog boxes.
    \item \textbf{Limitations in exploration and adaptability:} Particularly for agents equipped with modules like~\enquote{Set-of-Mark} (SoM), it has been observed that they can constrain the agent's action space, hindering exploration and adaptability to diverse tasks.
    \item \textbf{Significant performance gap with human capabilities:} As reported on the OSworld website~\cite{osworldWeb}, humans achieve a task completion rate of more than 72.36\%. In contrast, leading models reach approximately 42.9\% completion (as of June 2025), indicating a substantial gap with human performance.

\end{enumerate}
To address these challenges and guide the investigation of agent design, this research presents a set of questions to explore the architectural components, integration strategies, and generalization capabilities of LLM-based agents.

\subsection{Research Questions}
To guide this survey, we formulate the following research questions that structure the analysis of architectural foundations, subsystem design, and evaluation of LLM based agents.

\begin{enumerate}
    \setlength{\itemsep}{0.3em}
    \item \textbf{RQ1, Design space,} What architectural options exist for the core subsystems of LLM-based agents, perception, reasoning and planning, memory, and execution, and how can they be systematically organized for practitioner use?
    \item \textbf{RQ2, Integration,} Which subsystem integration patterns enable reliable closed-loop autonomy in realistic software environments, for example, GUI and web tasks that combine visual grounding with structured signals such as DOM or accessibility trees~\cite{wang2024oscar,kil2024dualview}?
    \item \textbf{RQ3, Reasoning efficacy,} How do reasoning strategies, for example, CoT, ToT, ReAct, and parallel planning, such as DPPM or MCTS-based approaches, affect task success rate, efficiency, and cost?
    \item \textbf{RQ4, Memory impact,} How do long-term and short-term memory mechanisms, for example, RAG and context management, influence accuracy, robustness to context length limits, and adaptation in long-horizon tasks?
    \item \textbf{RQ5, Failures and mitigation,} What are the principal failure modes in agentic settings, for example, hallucination, GUI misgrounding, repetitive loops, and tool misuse, and which mitigation techniques, for example, reflection, anticipatory reflection, SoM, and guardrails, are most effective?
    \item \textbf{RQ6, Evaluation and generalization,} Which benchmarks and metrics are appropriate for assessing these systems, for example, OSWorld, WebArena, and Mind2Web~\cite{osworld2024,zhou2024webarena,deng2023mind2web}, and to what extent do agents generalize across tasks, applications, and interfaces?
\end{enumerate}

Before delving into these research questions, let us first explore the origins of LLM-based agents.

\section{Fundamentals}
\subsection{Background of LLMs}
The introduction of machine learning methods, particularly deep learning, brought a significant shift by laying the groundwork for advanced modern AI models. 
Large language models (LLMs) are among the most significant developments.
Their appearance represents a major breakthrough in AI's ability to understand and produce complex language, influencing the state of LLM-based agents today and their future course.

A key technological advance in the development of LLMs has been the transformer architecture, distinguished by its~\enquote{attention mechanism}~\cite{vaswani2017attention}. This mechanism allows LLMs to attend to different words in the input enabling them to understand long-range dependencies~\cite{vaswani2017attention}. This architectural shift, alongside their training on vast datasets and the principles of generative AI, has enabled LLMs to perform a wide range of tasks, including natural language processing (NLP), machine translation, vision applications, and question-answering.

\subsection{From LLMs to LLM Agents}
LLMs in their standard form have significant limitations due to their chatbot nature. This restricts their effectiveness in real-world tasks. These models lack long-term memory, cannot autonomously interact with external tools, and struggle to pursue goals in dynamic environments. Such shortcomings hinder their performance in scenarios requiring sustained reasoning or multi-step workflows~\cite{xi2023risepotential}.

To overcome these constraints, LLMs are guided to follow a reasoning path and are provided with tools to interact with the environment that enables them to function as autonomous agents. They are well-suited for dynamic tasks because they exhibit good planning skills, context adaptability, and they minimize human intervention. Such agents offer a scalable and flexible solution by simulating human-like team strategies and leveraging external tools~\cite{jin2024llms}.
\smallskip

However, simply augmenting an LLM with modules, tools, or predefined steps does not make it an agent, in any case, that would make it a workflow.

\subsection{Workflows vs. Agents}
Many people confuse workflows with agents, but while both enhance the capabilities of large language models (LLMs), they are fundamentally different. Workflows are structured systems that enhance LLMs by enabling tool use, environmental interaction, or access to long-term memory. However, they are not agents. Workflows perform well in controlled and predictable environments where tasks are well defined and follow a fixed sequence of steps. In a workflow, the LLM follows a pre-established plan created by its designer, broken down into specific, sequential actions. This rigidity makes workflows highly effective for repetitive and structured tasks but limits their adaptability. If, during the workflow, the LLM faces an error, it often struggle to adjust, as they lack the ability to dynamically re-plan or adapt based on new information.

In contrast, agents are far more versatile and autonomous. Agents are designed to act according to the feedback from its environment. Rather than relying on a pre-set plan, agents generate their own strategies tailored to the task and context, often using techniques like Chain-of-Thought reasoning or iterative refinement to break down complex problems. This adaptability allows agents to deal with unexpected challenges, bounce back from mistakes, and function well in unpredictable environments~\cite{AnthropicAgents}.
\smallskip

To understand how these agents achieve autonomy, we first explore their core components and their interconnections.
\subsection{Constitution of an Agent}

\subsubsection{Perception System}
An agent begins its interaction with the world through its perception system. This component is responsible for capturing and processing data from the environment, such as images, sounds, or any other form of information. Its task is to transform this information into meaningful representations that the LLM can understand and utilize, such as identifying objects or recognizing patterns.

\subsubsection{Reasoning System}
The reasoning system receives the task instructions along with the data from the perception system and formulates a plan that is broken down into distinct steps. It is also responsible for adjusting this plan based on environmental feedback and evaluating its own actions to correct errors or improve execution efficiency.

\subsubsection{Memory System}
The memory system keeps the knowledge that is not embedded in the model’s weights. This includes everything from past experiences to relevant documents and structured data stored in relational databases. The LLM uses this information to enhance the accuracy of its responses.

\subsubsection{Action System}
Finally, the action system is responsible for translating abstract decisions into concrete actions that impact the environment. This module ensures that the agent's instructions are carried out in the real or simulated world, completing the interaction cycle by executing what has been decided. 
This can involve using a set of tools, such as calling APIs or writing code to execute mouse movements in a software environment~\cite{mi2025building}.

\begin{figure}[!htbp]
\centerline{\includegraphics[scale=0.4]{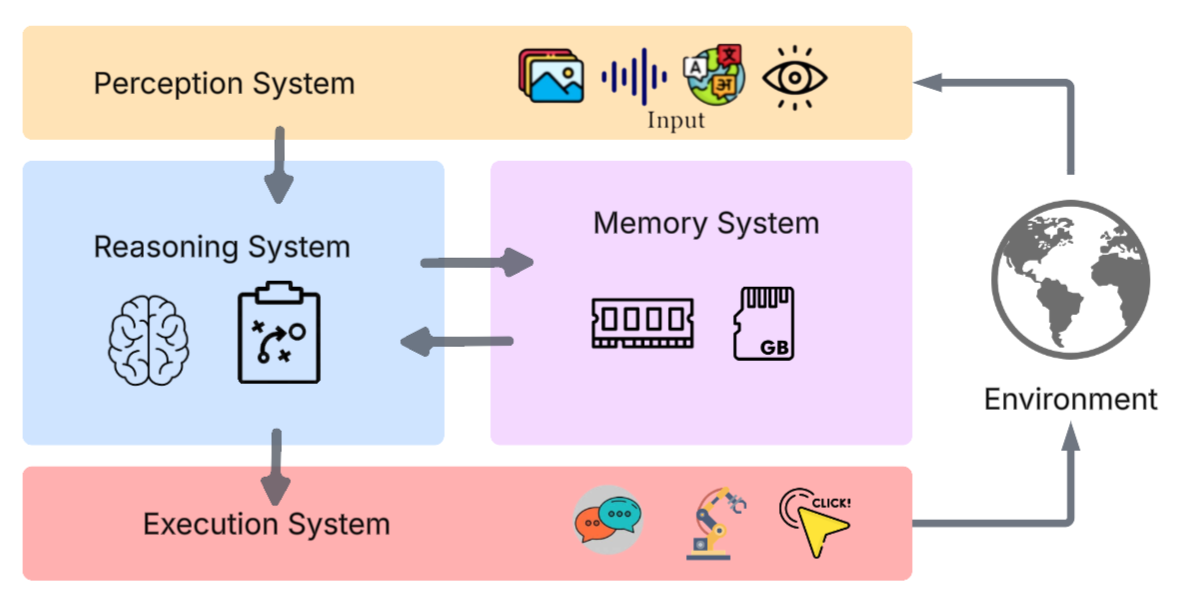}}
\caption{Key Components of an Agent's LLM Architecture}
\label{fig:openworld}
\end{figure}

Having outlined the core components that enable an LLM agent’s autonomy, we now delve into a detailed exploration of the perception system.

\section{Perception System}
The perception system of an LLM agent essentially acts as its~\enquote{eyes and ears,} converting environmental stimuli into a format that the LLM can understand and process. The complexity of the environment and the kinds of information required determine the architecture. This challenge can be approached in four ways: text-based, multimodal, information tree/structured data, and tool-based.

\subsection{Text-Based Perception (Pure LLM)}
The simplest form in which the environment is described is purely in text. The LLM receives and processes this text description. In this mode, the environment provides textual observations directly to the LLM's prompt. This could be a description of the current state, recent events, or results of actions taken. In this environment, the perception system does not need to intervene.

This approach offers low computational overhead for perception and integrates directly with the LLM's core capabilities. However, it is limited to environments that give the response to LLM interactions in text. This is practical for chats or text-driven simulations.

\subsection{Multimodal Perception}
Agents can process and integrate information from a variety of sources, mainly textual and visual (images, videos), thanks to multimodal perception. For agents functioning in real-world or graphical user interfaces (GUIs), this capability is crucial. In the context of LLM agents, this is largely achieved through Vision-Language Models (VLMs) and their more advanced successors, Multimodal Large Language Models (MM-LLMs). These models aim to bridge the gap between images and words, allowing agents to understand and generate content across both modalities.

Although significant progress has been made in the extension of LLMs to vision, it still has some challenges. For instance, most models still struggle with precise spatial relationships or accurate object counting without external aid~\cite{bordes2024introductionvisionlanguagemodeling}.

Regardless of the specific training paradigm, a fundamental principle is the learning of a unified embedding space for vision and language. This means that both visual and textual data are converted into numerical representations (embeddings) that can be processed and compared together by the model~\cite{li2025visuallargelanguagemodels}. 

MM-LLMs represent a significant advancement, distinguished by their approach of augmenting powerful, off-the-shelf LLMs to support multimodal inputs or outputs. Unlike VLMs, which primarily aim to align visual and linguistic representations, MM-LLMs leverage the inherent reasoning capabilities of a large language model as their central processing unit. This enables them not only to process and connect modalities but also to perform complex reasoning, planning, and generation across a diverse range of multimodal tasks.

The general architecture of MM-LLMs typically comprises a structured pipeline with distinct components~\cite{zhang2024mmllmsrecentadvancesmultimodal}:

\begin{itemize}
\setlength{\itemsep}{0.3em}
    \item \textbf{Modality Encoder (ME):} This component is responsible for encoding inputs from various modalities, such as images, videos, or even audio and 3D data, to obtain corresponding features or embeddings. For visual inputs, specialized encoders like Convolutional Neural Networks (CNNs) or Vision Transformers (ViT) are used to extract rich visual representations~\cite{li2025visuallargelanguagemodels,radford2021learningtransferablevisualmodels}.

    \item \textbf{Input Projector:} This component aligns the encoded features from non-textual modalities (e.g., visual embeddings) with the text feature space of the LLM. It acts as a bridge, transforming the visual embeddings into a format that the LLM can comprehend and integrate alongside textual inputs. This processing ensures that the visual embeddings are effectively supplied to the LLM, enabling the LLM to leverage its pre-trained linguistic knowledge for multimodal reasoning~\cite{li2025visuallargelanguagemodels,song2025bridgegap}.

    \item \textbf{LLM Backbone:} This is the core reasoning engine. The processed and aligned multimodal representations (visual embeddings and textual features) are fed to the LLM. The LLM processes these representations, answering using the semantic understanding of the inputs.

    \item \textbf{Output Projector (for multimodal generation):} For tasks requiring outputs in other modalities (e.g., generating images), this component maps signal token representations from the LLM Backbone into features understandable by a Modality Generator.

    \item \textbf{Modality Generator (for multimodal generation):} This component is tasked with producing outputs in distinct modalities, such as synthesizing images using models like Latent Diffusion Models.

\end{itemize}

\begin{figure}
    \centering
    \includegraphics[width=0.9\linewidth]{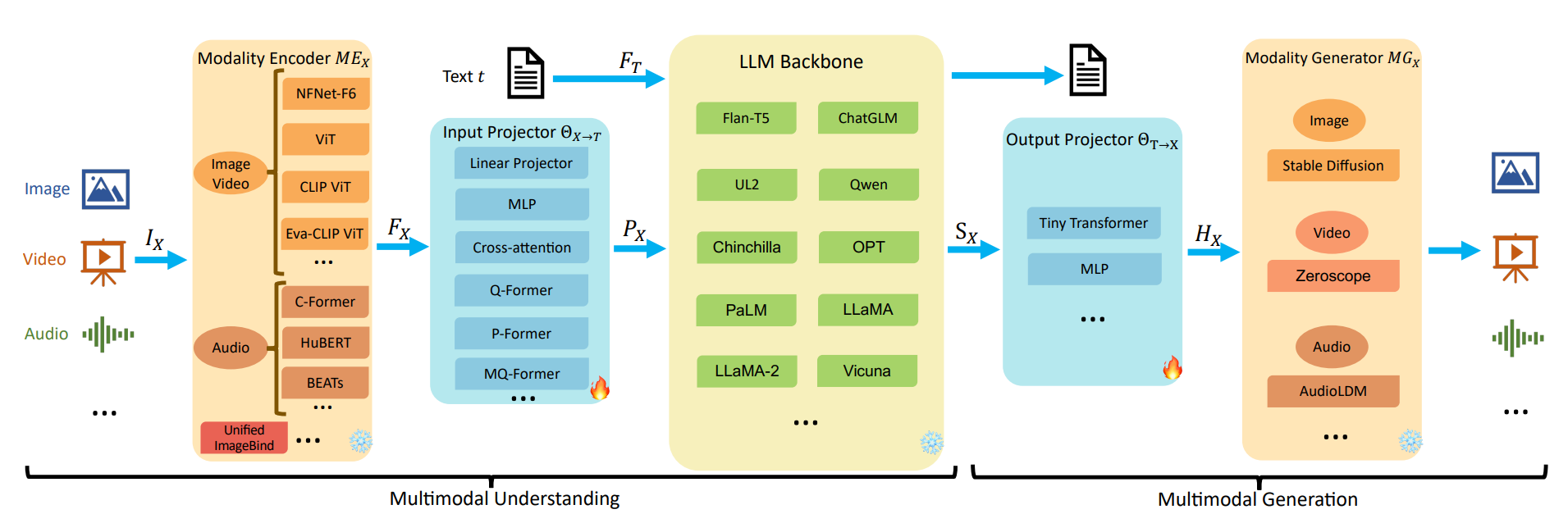}
    \caption{Architecture of Multimodal Large Language Models (MM-LLMs) for Understanding and Generation~\cite{zhang2024mmllmsrecentadvancesmultimodal}}
    \label{fig:enter-label}
\end{figure}

While the architectural components of MM-LLMs enable multimodal processing, their perceptual capabilities often require further enhancement to address limitations in visual understanding, as explored in the following subsection.

\subsubsection{Enhancing Perception in MM-LLMs}
As outlined in the paper~\enquote{VCoder: Versatile Vision Encoders for Multimodal Large Language Models} by Jain et al. (2023)~\cite{jain2023vcoder}, traditional MM-LLM systems often face limitations in fundamental visual perception, such as accurately identifying or counting objects, and a tendency to hallucinate non-existent entities. 

A faster and more cost-effective way to enhance perception (rather than improving each individual component of an MM-LLM) is to use visual encoders. These encoders, which can be separate models, extract relevant information from images to help the MM-LLM interpret them more effectively. 
While this approach doesn't match the performance gains of directly improving each component of the MM-LLM, it offers a practical trade-off by significantly improving results at a much lower computational and developmental cost. These are different ways to enhance visual perception with visual encoders:

\begin{itemize}
    \setlength{\itemsep}{1em}
    \item \textbf{Segmentation and Depth Maps:}
    VCoder enhances MM-LLM capabilities through a specialized adaptive architecture and the integration of additional perception modalities. It functions as an adapter to a base MM-LLM, enabling the model to process~\enquote{control inputs} such as segmentation maps (offering fine-grained object and background information) and depth maps (providing spatial relationship details). Information from these inputs is projected into the LLM's embedding space via additional vision encoders~\cite{radford2021learningtransferablevisualmodels}.

    \begin{figure}
        \centering
        \includegraphics[width=0.8\linewidth]{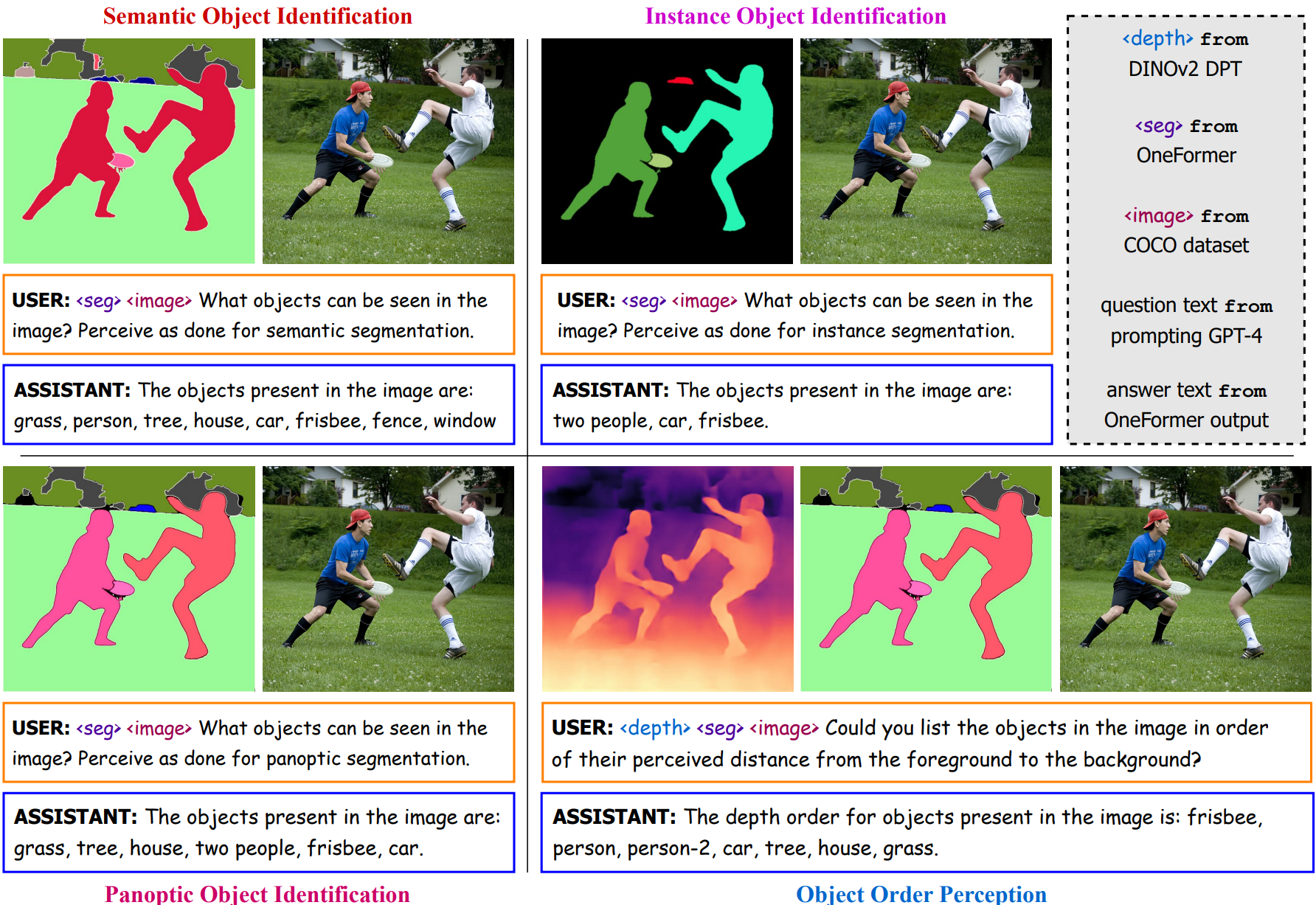}
        \caption{Usage of segmentation and depth maps for MM-LLM perception\cite{jain2023vcoder}}
        \label{fig:enter-label}
    \end{figure}

    \item \textbf{Set-of-Mark Operation:} To enhance the model's ability to handle complex visual tasks, Set-of-Mark (SoM) operation provides a structured approach to guide MM-LLMs in processing visual inputs. As seen in Fig.~\ref{fig:set-of-mark} set-of-mark process consists in annotating images with explicit markers (e.g., bounding boxes or labels) that highlight key regions or objects, enabling the model to focus on specific areas during reasoning. This technique improves the model's understanding of the image and task-specific performance~\cite{yang2023setofmark}.

    \begin{figure}
        \centering
        \includegraphics[width=1\linewidth]{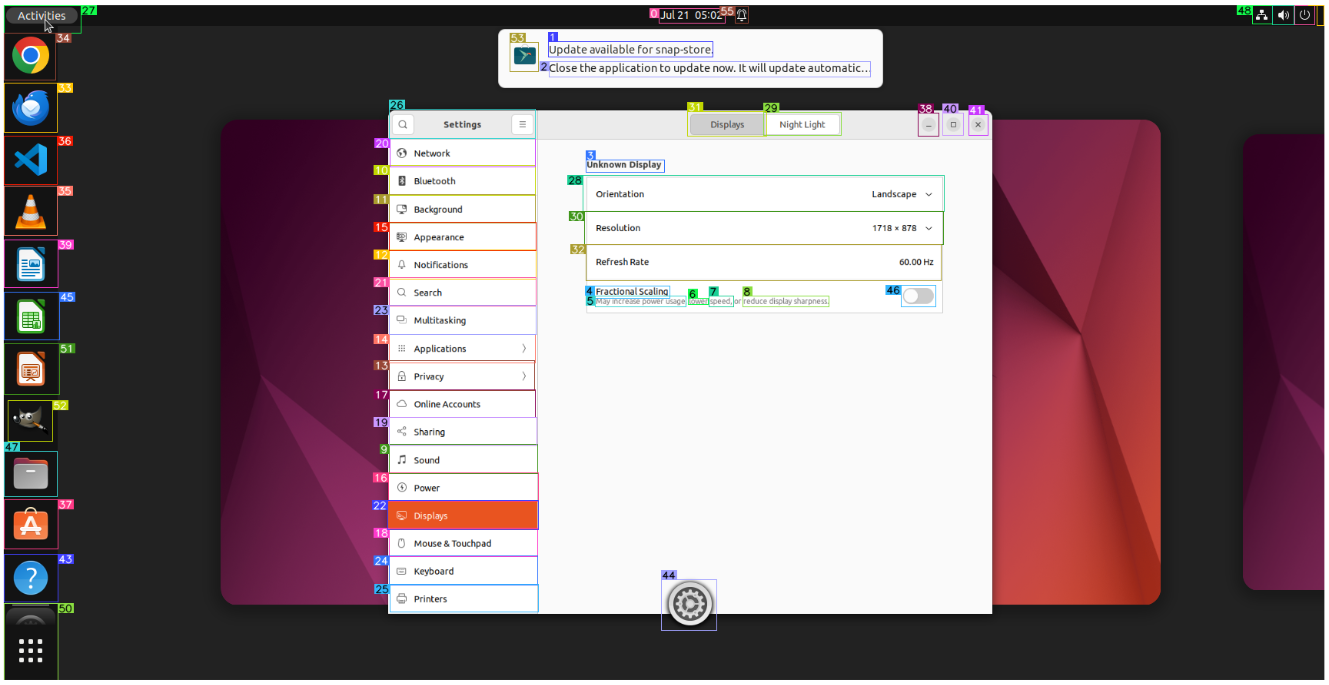}
        \caption{Image with Set-of-Mark~\cite{yang2023setofmark}}
        \label{fig:set-of-mark}
    \end{figure}

\end{itemize}

Experimental evidence presented in the papers~\cite{jain2023vcoder,yang2023setofmark} indicates that MM-LLMs adapted with VCoder and SoM significantly outperform baseline models on object-level perception tasks, demonstrating improved counting accuracy and reduced hallucination. This highlights the ongoing efforts to enhance the granular perception capabilities of LLM-based agents.

While techniques like Set-of-Mark and VCoder enhance visual perception through targeted annotations and prompting, structured data approaches, such as Accessibility Tree and HTML utilization, offer alternative methods for robust environmental interpretation, as explored in the following subsection.

\subsection{Information Tree/Structured Data Perception}
\begin{itemize}
    \item \textbf{Accessibility Tree Utilization:} OSCAR~\cite{wang2024oscar} utilizes an A11y tree generated by the Windows API for representing GUI components, incorporating descriptive labels to facilitate semantic grounding.
    \item \textbf{HTML Utilization:} Meanwhile, DUALVCR~\cite{kil2024dualview} captures both the visual features of the screenshot and the descriptions of associated HTML elements to obtain a robust representation of the visual screenshot.
\end{itemize}

\subsection{Tool-based Perception}
Beyond direct multimodal inputs and structured data retrieval, LLM-based agents can significantly enhance their perception capabilities through tool augmentation. This means utilizing external tools and APIs to enable the agent to gather, process, and interpret data from a wider variety of sources, including real-world sensors and specialized databases. The mechanism of integration typically involves the LLM generating specific tool calls based on its current understanding and goals, with the results from these tools being~\enquote{fed back} into the LLM~\cite{schick2023toolformer,yu2023gorilla}.

\subsubsection{Categorizing Tools for Perception}
The diverse landscape of external tools available to LLM agents can be broadly categorized based on the type of information they help perceive:

\begin{itemize}
    \setlength{\itemsep}{0.3em}
    \item \textbf{Web Search and Information Retrieval APIs:} These tools allow agents to access vast amounts of up-to-date information, facts, and specific data points from the internet. By issuing queries to search engines (e.g., Google Search API) or structured knowledge bases (e.g., Wikipedia API), agents can perceive real-time events, verify facts, or retrieve details beyond their training data cutoff. This helps the agent fill in missing environmental information and is crucial for tasks requiring current affairs knowledge or factual accuracy~\cite{schick2023toolformer,yu2023gorilla,nakano2022webgptbrowser}.
    
    \item \textbf{Specialized APIs:} Agents can use domain-specific APIs designed for specific data types. Examples include weather APIs (for perceiving current and forecasted climatic conditions), stock market APIs (for real-time financial data), or scientific databases and literature APIs (for accessing specialized research papers and experimental data). 
    These tools enable agents to perceive specific information relevant to niche tasks~\cite{yu2023gorilla,li2023apibank}, and can be implemented as document-centric microservices for knowledge discovery~\cite{gidey2022document}.
    
    \item \textbf{Sensor Integration (Conceptual via Intermediary Tools):} While an LLM agent does not directly interface with physical hardware sensors, its perception system can be augmented to interpret data originating from them. This is achieved through intermediary tools or services that convert raw sensory data (e.g., temperature readings, GPS coordinates, accelerometer data) from real-world or simulated environments into a digestible format (textual descriptions, structured data like JSON). This allows the agent to perceive physical properties and spatial relationships of its environment, crucial for tasks in robotics or interactive simulations~\cite{chen2024robogpt,brohan2023rt2}.
    
    \item \textbf{Code Execution Tools:} These tools enable agents to execute code for data processing and calculations. By generating and executing code (e.g., Python scripts via an interpreter), agents can perceive insights from raw data, such as parsing complex log files, running statistical analyses on datasets, or querying local databases. This allows for dynamic and flexible data interpretation beyond simple text matching~\cite{openai2023codeinterpreter,gao2023pal}.
    
\end{itemize}

Let’s now explore how integrating the diverse perception system approaches empowers an LLM agent to effectively handle tasks, as illustrated in a practical example.

\subsection{Example of a Perception System in an LLM Agent}
Let’s consider an LLM agent designed to automate tasks within a Graphical User Interface (GUI), such as managing emails in a web-based application.

Although this could be easier to achieve using the email API, imagine a scenario where the agent's objective is to identify, classify, and, if necessary, respond to incoming company emails.

To achieve this, the agent starts by capturing a screenshot of the email app. It then applies a Set-of-Mark operation using a visual encoder. This encoder draws a box on every interactive element on the screen, such as buttons or checkboxes and stores the coordinates of each box. The output consists of the image with the bounding boxes and a structured list describing each detected element, including its text content (if any), a brief description, and its coordinates.

In parallel, the agent retrieves the Accessibility Tree (A11y Tree) or the HTML source of the page~\cite{gidey2025affordance}. This tree provides a hierarchical representation of GUI components, such as buttons, text fields, links, and list items—along with their roles, labels, states (e.g.,~\enquote{unread}). Such data is typically extracted through browser automation tools.

The accessibility tree and the visual encoder output combine to create a perception system. This system allows the agent to understand the interface: its visual layout, the semantics and roles of individual elements, and their spatial structure. When combined with the image understanding capabilities of a MM-LLM, this perception system enables the agent to build a rich, actionable model of the GUI environment.

Despite the robustness of this perception system, it has a number of drawbacks and restrictions that can impact its performance and reliability.

\subsection{Perception Challenges and Limitations}

While significant progress has been made in empowering LLM agents with advanced perceptual capabilities, several critical challenges and limitations persist across all approaches:

\begin{itemize}
\setlength{\itemsep}{0.3em}
    \item \textbf{Hallucination:} The tendency for models to~\enquote{hallucinate} non-existent objects or misinterpret visual cues remains a significant hurdle. This can lead to agents making decisions based on incorrect interpretations, resulting in errors or undesirable behavior\cite{huang2025}.
    
    \item \textbf{Latency in Inference Pipelines:} Integrating complex perception modules, especially those involving multimodal processing or external tool calls, can introduce substantial latency. Real-world applications, particularly those requiring real-time interaction (e.g., robotics, dynamic GUI automation), demand rapid perceptual updates. The sequential nature of many perception pipelines, from raw data acquisition to final LLM interpretation, can create bottlenecks, hindering the agent's responsiveness.
    
    \item \textbf{Context Window Limits:} Large inputs, such as high-resolution images or extensive structured data, can generate a vast amount of tokens or embeddings. Encoding and feeding this entire information into the LLM's context window can quickly exceed its limitations~\cite{wang2024beyondthelimits}.
    
    \item \textbf{Data Collection:} Training robust perception systems, particularly for multimodal or specialized domains, often requires large volumes of high-quality, annotated data. The collection of this data can be costly and time-consuming.
    
    \item \textbf{Computational Resources:} High-fidelity perception, especially with multimodal inputs, requires high computational resources for both training and inference. This can be a barrier for execution in resource-constrained environments or for widespread adoption.
    
\end{itemize}

Ultimately, the quality and fidelity of an LLM agent's perception system directly affects the reasoning and planning modules. Therefore, continuous advancements in perception technologies are not merely improvements to one component, but fundamental enablers for building more intelligent, reliable, and capable LLM agents.

\begin{sidewaystable*}[htbp]
\centering
\caption{Summary of Perception Approaches for LLM-Based Agents}
\begin{tabular}{>{\raggedright\arraybackslash}p{2.5cm} >{\raggedright\arraybackslash}p{2.5cm} >{\raggedright\arraybackslash}p{3.5cm} >{\raggedright\arraybackslash}p{4cm} >
{\raggedright\arraybackslash}p{4cm}}
\toprule
 \textbf{Modality} & \textbf{Input Format} & \textbf{Tool Dependencies} & \textbf{Strengths} & \textbf{Limitations} \\
\midrule
Text-Based Perception & Plain text descriptions & None (relies on LLM's native text processing) & Low computational overhead; seamless integration with LLM; ideal for text-driven environments & Limited to text-only environments; cannot process visual or other non-textual data \\
\addlinespace
Multimodal Perception & Text, image/video embeddings, audio transcripts & Vision-Language Models (e.g., CLIP, ViT), Multimodal LLMs, preprocessing tools (e.g., CNNs, ASR) & Processes diverse data types; suitable for GUIs and real-world tasks; leverages advanced VLMs & High computational cost, struggles with precise spatial tasks and requires extensive training data \\
\addlinespace
Information Tree/Structured Data Perception & JSON, XML, database records, A11y trees & Parsers, database query tools, accessibility frameworks & Precise semantic understanding; efficient for structured environments like GUIs or databases & Limited to environments with structured data and requires predefined schemas or parsing logic \\
\addlinespace
Tool-Augmented Perception & Tool outputs (text, JSON, numerical data) & External APIs, code interpreters, sensor interfaces, web search tools & Extends perception to real-time and specialized data; highly flexible and dynamic & Dependent on tool availability and reliability, complex integration and error handling \\
\bottomrule
\end{tabular}
\label{tab:perception_comparison}
\end{sidewaystable*}

\clearpage
Having established how the perception system equips an LLM agent with a comprehensive understanding of the GUI environment, as summarized in the preceding table, the next critical component is the reasoning system. This system leverages the processed perceptual input to make informed decisions and execute complex tasks.

\section{Reasoning System}
\subsection{Task Decomposition}
A key tactic for helping LLM agents solve complicated problems is task decomposition. This strategy divides the problem into smaller and easier-to-manage subtasks. This approach, akin to the~\enquote{divide and conquer} algorithmic paradigm, simplifies the planning process. The procedure involves two main steps: first, the~\enquote{decompose} step, where the complex task is broken into a set of subtasks; and second, the~\enquote{subplan} step, where for each subtask a plan is formulated~\cite{huang2024understanding}. This systematic breakdown helps in navigating intricate real-world scenarios that would otherwise be challenging to address with a single-step planning process.

Current methodologies for task decomposition broadly fall into two categories: Decomposition first and Interleaved decomposition~\cite{huang2024understanding}. Decomposition first methods, as seen in systems like HuggingGPT~\cite{shen2023hugginggpt} and Plan-and-Solve~\cite{wang2023plan}, initially decompose the entire task into sub-goals and then proceed to plan for each sub-goal sequentially. HuggingGPT, for instance, explicitly instructs the LLM to break down multimodal tasks and define dependencies between subtasks~\cite{shen2023hugginggpt}. 
A slightly modified version of the Decomposition first approach is DPPM (Decompose, Plan in Parallel, and Merge). It addresses the limitations of existing planning methods, such as:
\begin{enumerate}
    \item Handling heavy constraints
    \item Carrying errors from the planning of previous steps
    \item Forgetting the main goal
    \item Cohesion between subtasks
\end{enumerate}
DPPM tackles these problems with the following methods:
First, it decomposes the complex task into subtasks. Second, it generates subplans for each of these subtasks concurrently using individual LLM agents. This parallel planning allows each agent to focus only on its assigned subtask, promoting independent work and avoiding the cascading errors that can occur when subplans are sequentially dependent. Finally, DPPM merges these independently generated local subplans into a coherent global plan~\cite{lu2025decomposeplanparallel}. Although this method can struggle to adapt well to unexpected environmental problems, this limitation can be mitigated by reflecting on the plan after each execution step.

In contrast, interleaved decomposition methods, such as Chain-of-Thought (CoT)~\cite{wei2023chainofthought} and ReAct~\cite{yao2023react}, interleave the decomposition and subtask planning process, revealing only one or two subtasks at a time based on the current state. This dynamic adjustment based on environmental feedback enhances fault tolerance, although excessively long trajectories in complex tasks can sometimes lead to hallucinations or deviation from original goals~\cite{huang2024understanding}.

Further advancements in task decomposition and planning strategies include approaches such as RePrompting and ReWOO. RePrompting involves checking if each step of a plan meets necessary prerequisites before execution. If a step fails due to unmet prerequisites, a precondition error message is introduced, prompting the LLM to regenerate the plan with corrective actions~\cite{raman2023planning}. ReWOO introduces a modular paradigm that decouples reasoning from external observations, where agents first generate comprehensive plans and obtain observations independently, then combine them to derive final results~\cite{xu2023rewoo}.

\begin{figure}
    \centering
    \includegraphics[width=1\linewidth]{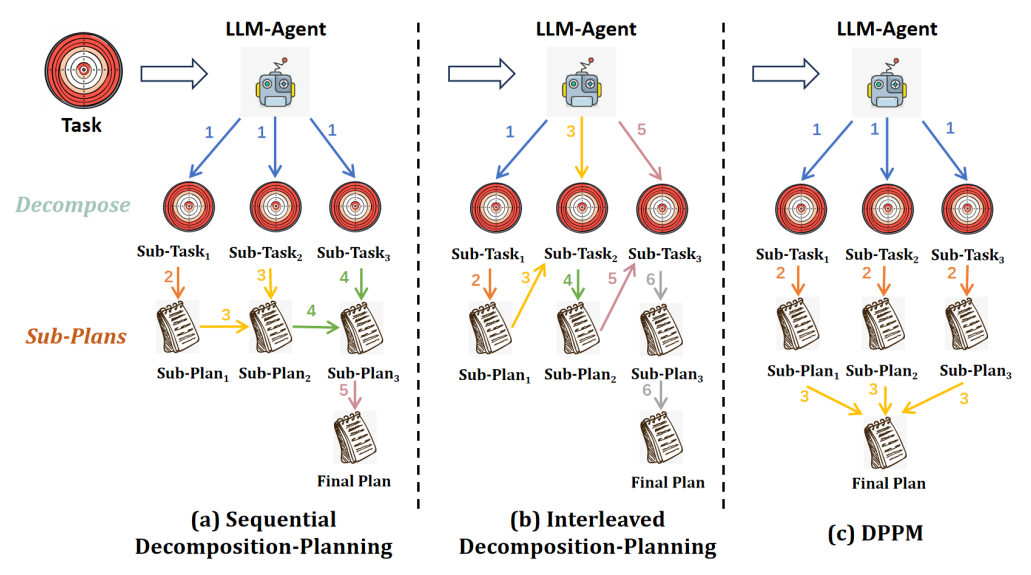}
    \caption{Comparison of different types of planning frameworks, including sequential decomposition-planning, interleaved decomposition-planning, and DPPM~\cite{lu2025decomposeplanparallel}.}
    \label{fig:enter-label}
\end{figure}

\subsection{Multi-Plan Generation and Selection}
Due to the inherent complexity of tasks and the uncertainty associated with LLMs, a single plan generated by an LLM Agent may often be suboptimal or even infeasible. To address this, multi-plan selection emerges as a more robust approach, focusing on leading the LLM to explore multiple alternative plans for a given task~\cite{wang2023selfconsistency}.
This methodology involves two main stages: multi-plan generation and optimal plan selection~\cite{huang2024understanding}. Multi-plan generation aims to create a diverse set of candidate plans, often by leveraging the uncertainty in the decoding process of generative models.

There are various strategies:
\begin{itemize}
\setlength{\itemsep}{0.3em}
    \item \textbf{Self-consistent CoT (CoT-SC):} This approach generates various reasoning paths and their corresponding answers using Chain of Thought (CoT), then selects the answer with the highest frequency as the final output~\cite{wang2023selfconsistency}.  
    \item \textbf{Tree-of-Thought (ToT) and Graph of Thoughts (GoT):} ToT generates plans using a tree-like reasoning structure where each node represents an intermediate~\enquote{thought.} The selection of these steps is based on LLM evaluations. Unlike CoT-SC, ToT queries LLMs for each reasoning step~\cite{yao2023tree}. Graph-of-Thought (GoT) extends the tree-like reasoning structure of ToT to graph structures.
    It supports arbitrary thought aggregation and allows for transformations of thoughts, leading to more powerful prompting strategies~\cite{besta2023graph}.

    \begin{figure}
        \centering
        \includegraphics[width=0.9\linewidth]{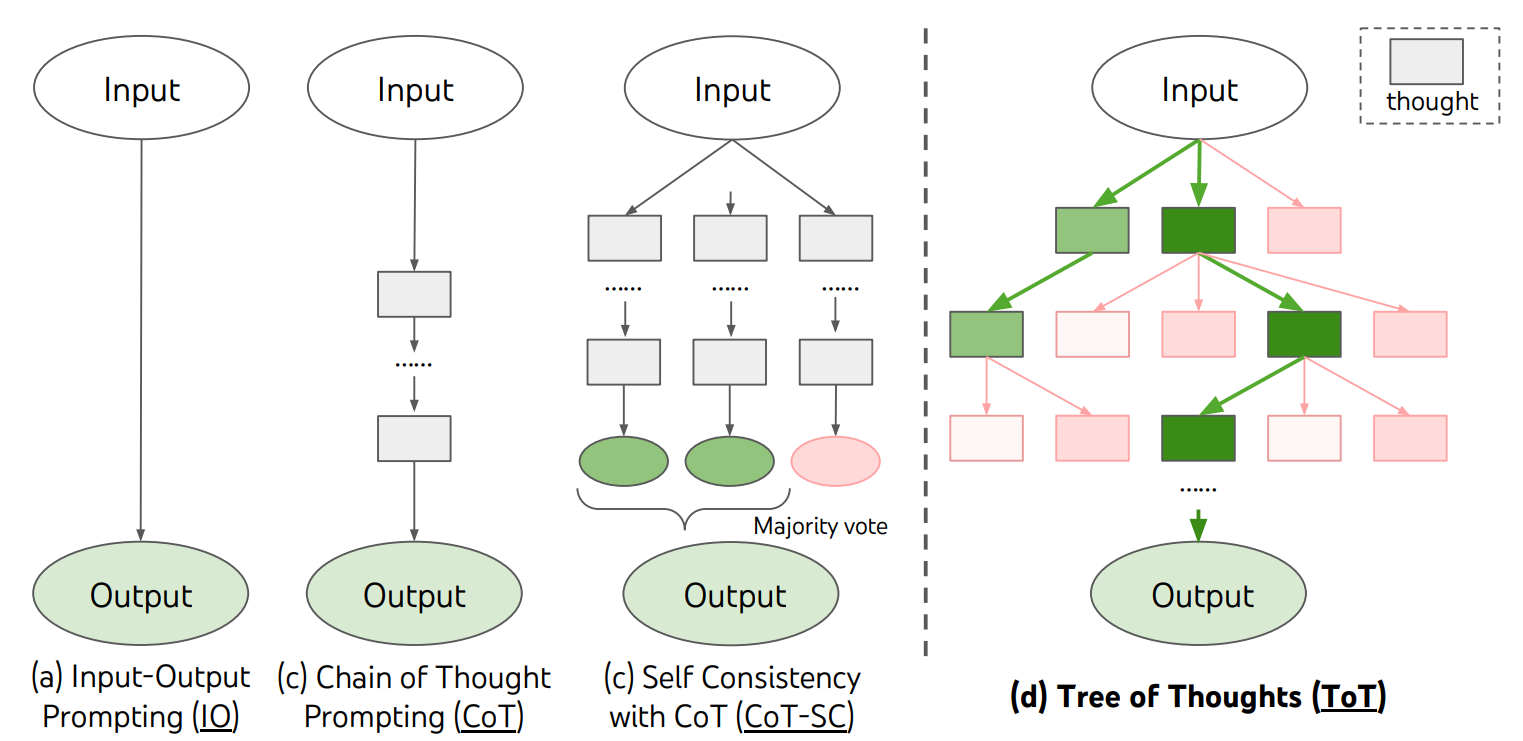}
        \caption{Schematic illustrating various approaches to problem solving with LLMs~\cite{yao2023tree}.}
        \label{fig:enter-label}
    \end{figure}
    
    \item \textbf{LLM-MCTS and RAP:}
    These methods leverage LLMs as a heuristic policy function for the Monte Carlo Tree Search (MCTS).
    Multiple potential actions (or plans) are obtained through multiple calls to the LLM during the MCTS process~\cite{zhao2023llm}. RAP~\cite{hao2023reasoning} specifically builds a world model to simulate potential benefits of different plans using MCTS to generate the final plan.
\end{itemize}

 Once a set of candidate plans is generated, the next step is plan selection, where different search algorithms are employed~\cite{huang2024understanding}. Self-consistency, for instance, utilizes a simple majority vote strategy to identify the most suitable plan~\cite{wang2023selfconsistency}. More advanced methods like Tree-of-Thought leverage tree search algorithms such as conventional Breadth-First Search (BFS) and Depth-First Search (DFS) for expansion and selection, evaluating multiple actions to choose the optimal one~\cite{yao2023tree}. Similarly, LLM-MCTS and RAP adopt tree structures to facilitate multi-plan searches using the MCTS algorithm~\cite{hao2023reasoning}. The scalability of multi-plan selection is a significant advantage, allowing for a broader exploration of solutions within expansive search spaces. However, this comes with trade-offs like increased computational demands. Furthermore, the reliance on LLMs for plan evaluation introduces challenges regarding their performance in ranking tasks and the potential for randomness due to the stochastic nature of LLMs, which can affect the consistency and reliability of chosen plans~\cite{huang2024understanding}.
 \smallskip

While multi-plan selection enables LLM agents to explore and evaluate multiple potential solutions prior to execution, the reasoning system is further enhanced by the process of reflection. This mechanism allows agents to evaluate their actions and outcomes after the execution, encouraging continuous improvement and adaptability in dynamic environments.

\subsection{Reflection}
Reflection, in the context of LLM agents, refers to the agent's ability to critically evaluate its own past actions, reasoning, and outcomes, and then use these insights to improve its future performance. This allows agents to learn from their mistakes or inefficiencies without human intervention. 

Key characteristics of reflection include:
\begin{itemize}
\setlength{\itemsep}{0.3em}
    \item \textbf{Self-Evaluation:} The agent examines its completed (or ongoing) task, its generated plans, and the results of its actions. This often involves comparing actual and expected outcomes.
    \item \textbf{Error Detection and Analysis:} Identifying where things went wrong, why a plan failed, or where the reasoning failed. This can be due to misunderstandings of the prompt, incorrect tool usage, logical inconsistencies, or environmental changes. Papers like~\cite{shinn2023reflexion} and~\cite{madaan2023selfrefine} exemplify this capability, where agents analyze their own outputs or execution traces to pinpoint issues.
    \item \textbf{Correction and Improvement:} Based on the analysis, the agent generates actionable insights. This might involve modifying its planning strategy, correcting its reasoning process, learning better ways to use tools, updating its~\enquote{memory} or state~\cite{shinn2023reflexion}, or generating a revised plan or a new set of actions~\cite{chen2023teachingllm,madaan2023selfrefine}. 
    \item \textbf{Goal-Driven Reflection:} Agents can reflect not just on errors, but also on efficiency or completeness, aiming to optimize their path to the goal even if no explicit error occurred.
\end{itemize}

Building on the conceptual framework of reflection and its key characteristics, we now explore the practical steps and components required to implement an effective reflection system in LLM agents.

\subsubsection{How to Implement a Reflection System:}
A Reflection system, as described in the paper~\enquote{Reflection: Language Agents with Verbal Reinforcement Learning,}~\cite{shinn2023reflexion} is a framework designed to improve the performance of language agents through linguistic feedback rather than traditional weight updates. It operates iteratively, allowing an agent to learn from its past mistakes by writing the feedback and storing and using these reflections in the next iterations. 
Here's a brief explanation of how to implement such a system:

Core Components:
\begin{itemize}
\setlength{\itemsep}{0.3em}
    \item \textbf{Actor:} This is typically a LLM that generates text and actions based on the current state observations and its memory.
    \item \textbf{Evaluator:} This component assesses the quality of the Actor's generated outputs. It takes a complete trajectory (sequence of actions and observations) and computes a reward score. Evaluation can be based on exact match grading, predefined heuristics, or even another LLM instance.
    \item \textbf{Self-Reflection Model:} Another LLM serves as the self-reflection model and is responsible for generating verbal self-reflections. Given a sparse reward signal (e.g., success/fail) and the current trajectory, it produces nuanced and specific feedback.
\end{itemize}

The paper~\enquote{DEVIL'S ADVOCATE: Anticipatory Reflection for LLM Agents}~\cite{wang2024devilsadvocate} introduces a distinct perspective: Anticipatory Reflection. This consists of the agent proactively reflecting on potential failures and considering alternative remedies before executing an action, essentially acting as a~\enquote{devil's advocate} to challenge its own proposed steps. This front-loaded introspection enhances consistency and adaptability by allowing the agent to anticipate and mitigate challenges, improving its ability to navigate complex tasks effectively.

 \subsection{Example of a Reasoning System}
A reasoning system can be developed by integrating some of the features mentioned above. Its core mechanism could be DPPM (Decompose, Plan in Parallel, and Merge).

First, the agent would decompose the main task into smaller subtasks. Then, in separate calls to an LLM, different planning options would be generated for each subtask. While generating these options, the LLM would consider potential issues that might arise during the execution of each subtask. Based on these anticipated problems, it would propose alternative approaches to either solve or avoid them. This process combines ideas from Tree-of-thought and the Anticipatory Reflection of the~\enquote{DEVIL'S ADVOCATE} paper mentioned before. 

Following the Merge step in DPPM, the agent would integrate the different subtask plans into a final, coherent plan to accomplish the overall goal. To do this, it would explore various combinations of the subtask options, ensuring that the resulting plan is logically consistent and that all subplans contribute meaningfully toward completing the main task.

After the final plan is constructed, it would be divided into groups of executable steps. As the agent carries out each group of steps, it would receive feedback from the environment. This feedback would be processed by a reflection mechanism, which would determine the current scenario:

\begin{enumerate}
    \item Successful execution: The actions produced the expected result, so the agent continues with the next group of steps.

    \item Minor error: The actions were close but not entirely accurate (e.g., the agent missed clicking a button because the coordinates were slightly off). In this case, the steps would be adjusted and corrected accordingly.

    \item Execution failure: The plan cannot be completed as-is (e.g., the button to be clicked does not exist). Here, the agent must reflect on whether the issue lies within the specific subplan or if the entire plan needs to be reconsidered. If only the subplan is flawed, a new one would be generated. If the problem is more fundamental, the entire planning process would restart from the beginning.
\end{enumerate}

 \begin{figure}
     \centering
     \includegraphics[width=0.9\linewidth]{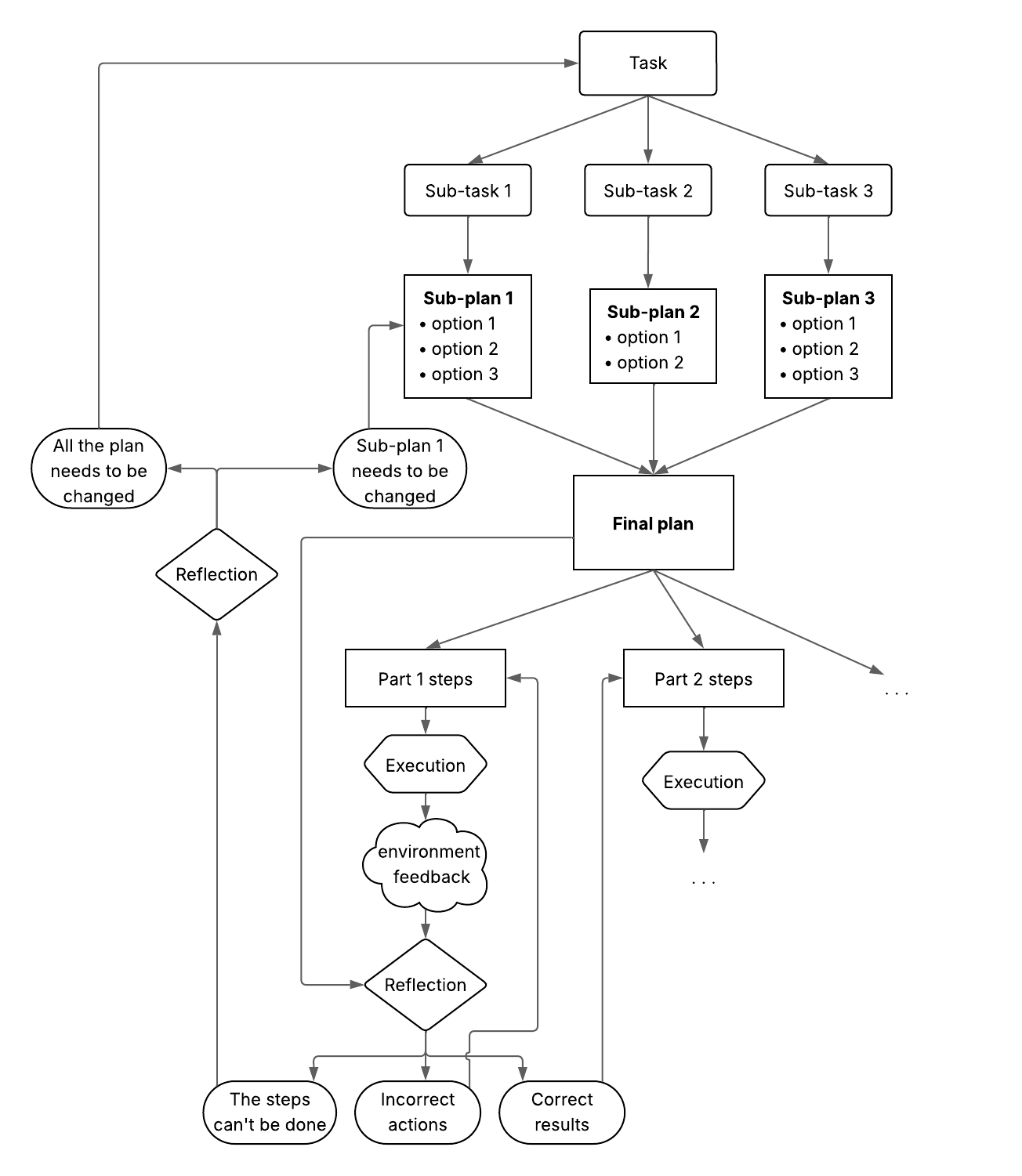}
     \caption{Flowchart of a Reasoning System Using Decompose, Plan, and Merge (DPPM) approach with a reflection system}
     \label{fig:enter-label}
 \end{figure}
\clearpage

Having illustrated how a single LLM agent can leverage a reasoning system like DPPM, combined with reflection, we now explore how multi-agent systems distribute these processes across specialized components to achieve greater scalability and efficiency.

\subsection{Multi-Agent Systems}
Expanding on the idea of multi-agent systems, a single agent can be made up of different specialized~\enquote{experts,} each of whom focuses on a distinct aspect of the interaction or reasoning. This modularity enables specialization at each step, increasing its capabilities and robustness~\cite{cai2025survey}.
Here are some examples of such useful experts that an LLM agent could integrate:
\begin{itemize}
\setlength{\itemsep}{0.3em}
    \item \textbf{Planning Expert:} This expert focuses on strategic thinking and task decomposition. Its role is to break down complex objectives into a series of manageable subtasks. This aligns with the actor component discussed in the reflection system, where agents perform reasoning and planning to undertake complex tasks~\cite{li2024survey}.
    \item \textbf{Reflection Expert:} It is dedicated to evaluating plans, responses, and overall performance. This aligns with the evaluator component discussed in the reflection system~\cite{li2024survey}.
    \item \textbf{Error Handling Expert:} Specifically focused on identifying, diagnosing, and suggesting recovery strategies for errors. 
    This expert could analyze logs, identify common failure patterns, and propose fixes. For example, it could propose to scroll down if an item is not found in a webpage~\cite{talebirad2023multiagentcollaboration}. It can also support self-healing behaviors in adaptive architectures~\cite{gidey2023modeling}.
    \item \textbf{Memory Management Expert:} Responsible for handling the agent's memory. This expert ensures that relevant information is retrieved efficiently and that the agent's context is maintained effectively, which is a critical challenge in LLM-based multi-agent systems~\cite{li2024survey,han2025llm}.
    \item \textbf{Action Expert:} This expert knows how to translate plans into concrete interactions with the environment. It's skilled in generating the necessary commands or API calls to interact with external tools, web interfaces, or other systems. For example, it is responsible for creating the move and click mouse movements in benchmarks like OSWorld.~\cite{osworld2024,guo2024llm,li2024survey}.
\end{itemize}

In addition to the experts mentioned above, there could be other helpful experts depending on the use case. 
For example, there could be a Coding Expert for generating, debugging, and optimizing code~\cite{talebirad2023multiagentcollaboration}; 
an Information Retrieval Expert for efficiently acquiring knowledge from external sources~\cite{li2024survey,guo2024llm}; 
a Human-Computer Interaction (HCI) Expert for optimizing user experience through adaptive and intuitive communication; 
a Constraint Satisfaction Expert for ensuring adherence to predefined rules, constraints, and assurances in various applications~\cite{guo2024llm}, 
who can also leverage existing model-driven verification tools~\cite{gidey2018factum,gidey2019modeling}; 
and a Security Expert for mitigating vulnerabilities, promoting secure practices, and monitoring risks in multi-agent interactions~\cite{talebirad2023multiagentcollaboration,guo2024llm}.

Having outlined some possible experts within multi-agent systems, we now turn to the practical process of designing and building these experts.

\subsection{How to Build an Expert}
Building an~\enquote{expert} within an LLM agent involves a combination of design principles and leveraging the capabilities of Large Language Models

Define the Expert's Role and Scope (Profile and Specialization).
The first step is to precisely define the~\enquote{distinctive attributes and roles}~\cite{talebirad2023multiagentcollaboration} of your expert. This involves:
\begin{itemize}
\setlength{\itemsep}{0.3em}
    \item \textbf{Clear Specialization:} What specific task, domain, or reasoning capability will this expert excel at? (e.g., planning, code generation, error handling).
    \item \textbf{Input and Output:} What kind of information does this expert take as input, and what kind of output does it produce?
    \item \textbf{Boundaries:} What are the limitations of its expertise? When should other experts be consulted or take over?~\cite{li2024survey}.
\end{itemize}

\subsubsection{Equip with Knowledge}
An expert's effectiveness hinges on its specialized knowledge. This can be achieved by:
\begin{itemize}
\setlength{\itemsep}{0.3em}
    \item \textbf{Targeted Prompting:} Crafting precise and detailed prompts to steer the LLM toward performing as the expert, incorporating specific prompting techniques such as Chain-of-Thought to enhance its reasoning process.
    \item \textbf{Fine-tuning (if applicable):} For highly specialized tasks, fine-tuning a base LLM on a dataset relevant to the expert's domain can enhance its performance.
    \item \textbf{External Knowledge Bases:} Integrating the expert with external tools or databases that provide specific, up-to-date, or proprietary knowledge relevant to its role~\cite{guo2024llm}.
    \item \textbf{Memory Integration:} The expert may have access to its memory (short-term context and long-term knowledge) which can store past experiences or knowledge relevant to its task~\cite{li2024survey,han2025llm}.
\end{itemize}

With the methodology for crafting specialized experts established, the following example illustrates how these components collaborate within a multi-agent framework.

\subsubsection{Example of a Multi-agent System}
First, the planning expert decomposes the main task into subplans. This expert is also responsible for avoiding infinite loops or repeated attempts if problems occur. Additionally, it collaborates with the constraint satisfaction expert to ensure that no constraints are violated during planning.

Next, the execution expert generates the specific actions to be performed in the environment. If any tools are required, it consults the tool expert to determine which tools to use and how to use them. If executable code is needed beyond basic actions, the coding expert is called upon to produce it.

Once actions are executed, feedback from the environment is received and processed by the reflection expert, which works together with the error handling expert to diagnose issues and propose solutions. Based on this diagnosis, the reflection expert decides how to proceed.

To improve its recommendations, the memory expert retrieves past experiences or successful workflows related to similar tasks. This knowledge is used to inform and enhance the next steps proposed to the planning or execution experts.

\begin{figure}
    \centering
    \includegraphics[width=0.5\linewidth]{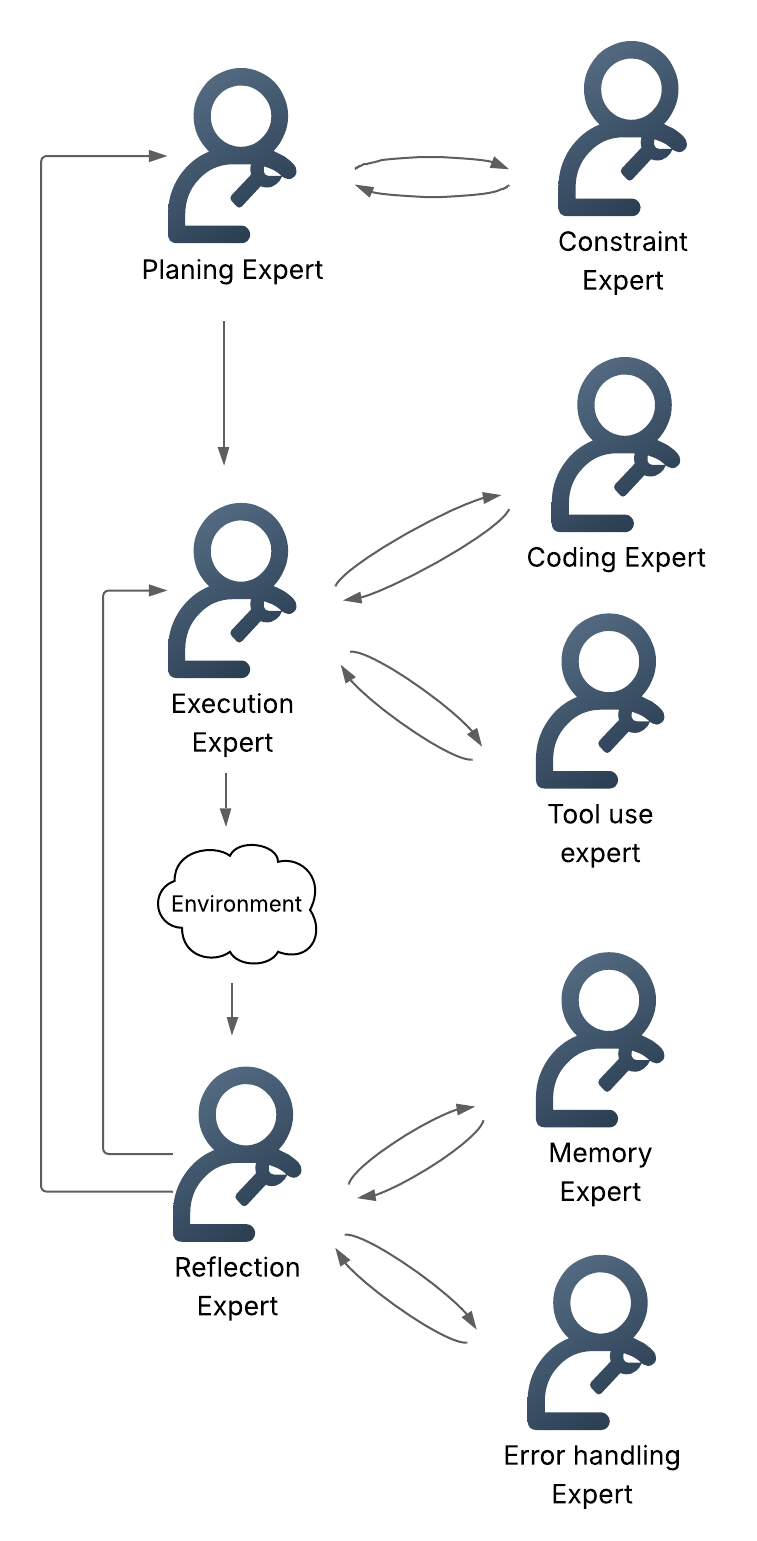}
    \caption{Example of the communication between agents in a multi-agent system}
    \label{fig:enter-label}
\end{figure}
\clearpage

\begin{sidewaystable*}[htbp]
\centering
\caption{Key Components and Techniques for the Reasoning System (Part 1)}
\begin{tabular}{>{\raggedright\arraybackslash}p{3cm} >{\raggedright\arraybackslash}p{3.2cm} >{\raggedright\arraybackslash}p{6cm} >{\raggedright\arraybackslash}p{3cm} >{\raggedright\arraybackslash}p{4cm}}
\toprule
\textbf{Component} & \textbf{Description} & \textbf{Key Techniques/\allowbreak Approaches} & \textbf{Advantages} & \textbf{Challenges/\allowbreak Limitations} \\
\midrule
\textbf{Task Decomposition} & Breaks down complex tasks into manageable subtasks to simplify planning and execution. & - \textbf{Sequential Decomposition}: Divides tasks into sequential subgoals and plans (e.g., Divide-and-Conquer). \newline - \textbf{Interleaved Decomposition}: Dynamically adjusts subtasks based on feedback (e.g., Chain-of-Thought [CoT], ReAct). \newline - \textbf{DPPM (Decompose, Plan in Parallel, Merge)}: Decomposes tasks, plans subtasks concurrently, and merges into a coherent global plan. & - Simplifies complex problem-solving. \newline - DPPM reduces cascading errors via parallel planning. \newline - Interleaved methods enhance fault tolerance. & - DPPM struggles with unexpected environmental changes. \newline - Interleaved methods may lead to hallucinations or deviation in long tasks. \\
\addlinespace
\textbf{Multi-Plan Generation and Selection} & Generates multiple candidate plans and selects the optimal one to address task uncertainty. & - \textbf{Self-consistent CoT (CoT-SC)}: Generates multiple reasoning paths and selects the most frequent answer. \newline - \textbf{Tree-of-Thought (ToT)}: Uses tree-like reasoning structures for plan generation. \newline - \textbf{Graph-of-Thoughts (GoT)}: Extends ToT with graph structures for flexible aggregation. \newline - \textbf{LLM-MCTS and RAP}: Use Monte Carlo Tree Search for plan generation and selection. & - Explores diverse solutions for robust planning. \newline - Scalable for complex tasks with large search spaces. & - High computational demands. \newline - Stochastic nature of LLMs may affect plan consistency. \newline - Challenges in ranking and evaluating plans. \\
\bottomrule
\end{tabular}
\label{tab:llm_agent_components_part1}
\end{sidewaystable*}

\clearpage

\begin{sidewaystable*}[htbp]
\centering
\caption{Key Components and Techniques for the Reasoning System (Part 2)}
\begin{tabular}{>{\raggedright\arraybackslash}p{3cm} >{\raggedright\arraybackslash}p{3,2cm} >{\raggedright\arraybackslash}p{6cm} >{\raggedright\arraybackslash}p{3cm} >{\raggedright\arraybackslash}p{4cm}}
\toprule
\textbf{Component} & \textbf{Description} & \textbf{Key Techniques/\allowbreak Approaches} & \textbf{Advantages} & \textbf{Challenges/Limitations} \\
\midrule
\textbf{Reflection} & Allows agents to evaluate actions post-execution, identify errors, and improve future performance. & - \textbf{Self-Evaluation}: Compares actual vs. expected outcomes. \newline - \textbf{Error Detection and Analysis}: Identifies and analyzes errors (e.g., incorrect tool usage, logical flaws). \newline - \textbf{Correction and Improvement}: Adjusts plans or strategies based on analysis. \newline - \textbf{Anticipatory Reflection (DEVIL'S ADVOCATE)}: Proactively considers potential failures before execution. & - Enables learning from mistakes without human intervention. \newline - Enhances adaptability and efficiency. \newline - Anticipatory reflection improves consistency. & - Requires robust feedback mechanisms. \newline - May be limited by the agent's ability to accurately self-evaluate. \\
\addlinespace
\textbf{Multi-Agent Systems} & Distributes reasoning tasks across specialized~\enquote{experts} for scalability and efficiency. & - \textbf{Planning Expert}: Handles task decomposition and strategic planning. \newline - \textbf{Reflection Expert}: Evaluates plans and suggests improvements. \newline - \textbf{Error Handling Expert}: Diagnoses and proposes fixes for runtime errors. \newline - \textbf{Others}: Includes Memory Management, Action, Coding, Information Retrieval, Dialogue Management, HCI, Constraint Satisfaction, and Security Experts. & - Enhances modularity and robustness. \newline - Leverages specialized expertise for complex tasks. \newline - Improves scalability through division of labor. & - Requires careful coordination between experts. \newline - Potential for increased complexity in system design. \newline - Security risks in multi-agent interactions. \\
\bottomrule
\end{tabular}
\label{tab:llm_agent_components_part2}
\end{sidewaystable*}

\clearpage
Having explored how reasoning systems enable LLM agents to plan, reflect, and collaborate on complex tasks, we now consider the memory system, which provides the critical foundation for retaining and applying past experiences to inform and enhance these reasoning processes.

\section{Memory System}
The memory system empowers LLM agents to manage information across varying time scales, with long-term memory anchoring sustained knowledge retention while short-term memory facilitates immediate contextual awareness.

\subsection{Long-term memory}
Long-term memory in LLM agents is crucial for sustained interaction and for the models to evolve and adapt over time. It allows agents to store relevant past memories and learn information from previous interactions. It also enables the agent to retain knowledge apart from its pre-trained knowledge. There are different ways of implementing it:

\begin{itemize}
\setlength{\itemsep}{0.3em}
    \item \textbf{Embodied Memory:}
    In the context of LLMs,~\enquote{embodied memory} often refers to the idea that an agent's experiences and learned behaviors become ingrained directly within its model parameters (weights) through continuous learning processes like fine-tuning. Unlike external memory systems, this type of memory is build into the model itself. When an LLM is fine-tuned on new data, it adjusts its weights, effectively encoding new~\enquote{facts} or~\enquote{experiences} directly into its neural network. This causes the model to act in ways similar to what it has learned from these experiences~\cite{xiang2023language}. 
    \item \textbf{RAG:}
    Retrieval-Augmented Generation (RAG) is a technique that enhances LLMs by using external knowledge to improve the accuracy of its responses. It operates in two main phases: retrieval and augmentation. Using a query, a retriever component first looks through an external knowledge base (often indexed by vector embeddings) to locate relevant documents. This gives the LLM access to updated and precise information that might not be encoded in its training data or within its immediate context window.
    
    Once the relevant information is retrieved, it is added to the LLM context alongside the original query. This augmented input enables the LLM to generate responses that are based on company files or personal documents making the response precise for the specific use case and reducing the likelihood of~\enquote{hallucinations}~\cite{lewis2021retrieval}.
    
    \item \textbf{SQL Database:}
    SQL databases are used to store structured knowledge, such as information about employees, orders, or other data that can be stored in a table. By converting natural language queries into SQL queries, text-to-SQL techniques facilitate reliable database interaction. Transformer-based models are especially well-suited for producing intricate SQL queries because of their attention mechanism~\cite{zhu2024llm}.

\end{itemize}

\subsection{Short-term memory}
Short-term memory in LLM agents is analogous to the input information maintained within the context window, which acts as a temporary workspace~\cite{wang2025survey}.
\smallskip

Regardless of whether it's for long-term retention or immediate contextual awareness, the memory module's effectiveness hinges on what kind of data to store.

\subsection{What Kind of Data to Store}
The memory module within an LLM agent's architecture is designed to store diverse types of information perceived from its environment and interactions. This stored data is then used to make better decisions, enabling the agent to accumulate experiences, evolve, and behave in a more consistent and effective manner.
\begin{itemize}
\setlength{\itemsep}{0.3em}
    \item \textbf{Experiences:} It is beneficial to store records of both successful and failed tasks. Research has indicated that even failed experiences, when appropriately logged and distinguished as such, can be valuable. By explicitly noting a~\enquote{failed experience,} LLMs can learn to avoid repeating similar mistakes in the future. This continuous learning from past interactions, including the identification of~\enquote{invalid action filtering,} contributes to the agent's robust development and ability to adapt~\cite{alazraki2025needexplanationsllm,hamdan2025llmslearnnegativeexamples}.
    To store an experience, you capture a task's natural language instruction (e.g.,~\enquote{Who ordered order 0130?}) and the sequence of steps taken to solve it, where each step includes the agent's observation of the environment (e.g.,~\enquote{The current page shows order 0130}) and the action performed (e.g., click(\enquote{126}) or stop()). This data, structured as an experience with the instruction and a trajectory of observation-action pairs, is saved in a storage system like a database or a JSON file within a collection of experiences. This format ensures that the experience is retrievable for later use, such as inducing a workflow with a summarized description and generalized steps, which can then be integrated into the agent’s memory to guide future tasks~\cite{wang2024agentworkflow}.
    \item \textbf{Procedures:} LLM agents can learn reusable task workflows from past experiences to guide future actions, similar to humans. Agent Workflow Memory (AWM) is a method that induces commonly reused routines (workflows) from training examples and then selectively provides these workflows to the agent to guide subsequent generations~\cite{wang2024agentworkflow}.
    \item \textbf{Knowledge:}
    This category encompasses external information received as facts, such as data from articles, company-specific information, details about machinery, and internal company rules~\cite{gao2024rag}, including document-based discovery pipelines in microservices architectures~\cite{gidey2022document}.

    \item \textbf{User information:}
    Beyond just user preferences, this includes personal information that the user has supplied, such as details about their past activities (e.g., where they spent the last Christmas) or background (e.g., where their parents are from). Mechanisms like MemoryBank aim to comprehend and adapt to a user's personality over time by synthesizing information from previous interactions, which inherently involves storing and utilizing these personal details~\cite{zhong2023memorybank}.
\end{itemize}

While defining what kind of data to store is crucial for an LLM agent's effectiveness, the utility and management of this stored information are inherently subject to several limitations.

\subsection{Limitations}
\begin{itemize}
\setlength{\itemsep}{0.3em}
    \item \textbf{Context Window:}
    Large Language Models (LLMs) operate with a fundamental constraint known as the~\enquote{context window} or~\enquote{context length.} This refers to the maximum amount of text (measured in~\enquote{tokens,} which can be words, parts of words, or punctuation) that an LLM can process and consider at any one time when generating a response or performing a task.
    The primary impact of a limited context window is that LLMs cannot directly integrate or utilize all information in very long sequences. 
    The easiest way to overcome this is to truncate large texts or summarize them~\cite{wang2024beyondthelimits}.
    \item \textbf{Memory Duplication:}
    When storing information in memory, a potential issue is handling data that is similar to existing records. Various methods have been developed to integrate new and previous records to address this~\enquote{Memory Duplication} problem. For instance, in one approach, successful action sequences related to the same sub-goal are stored in a list. Once this list reaches a size of five, all sequences within it are condensed into a unified plan solution using LLMs, and the original sequences are then replaced with this newly generated one. Another method aggregates duplicate information by accumulating counts, thereby avoiding redundant storage~\cite{wang2025survey}.
\end{itemize}

\begin{sidewaystable*}[htbp]
\centering
\caption{Memory Components for LLM-Based Agents (Part 1)}
\begin{tabular}{>{\raggedright\arraybackslash}p{3cm} >{\raggedright\arraybackslash}p{3cm} >{\raggedright\arraybackslash}p{6cm} >{\raggedright\arraybackslash}p{3cm} >{\raggedright\arraybackslash}p{4cm}}
\toprule
\textbf{Component} & \textbf{Description} & \textbf{Key Techniques/\allowbreak Approaches} & \textbf{Advantages} & \textbf{Challenges/Limitations} \\
\midrule
\textbf{Long-term Memory} & Stores knowledge for sustained retention, enabling agents to recall past experiences and synthesize information from previous interactions. & - \textbf{Embodied Memory}: Experiences are ingrained in the model's parameters through continuous learning (e.g., fine-tuning). \newline - \textbf{Retrieval-Augmented Generation (RAG)}: Retrieves relevant documents from an external knowledge base using vector embeddings to enhance responses. \newline - \textbf{SQL Database}: Stores structured data (e.g., employee or order details) accessible via text-to-SQL queries generated by LLMs. & - Enables persistent knowledge retention. \newline - RAG reduces hallucinations by grounding responses in verifiable sources. \newline - SQL databases support structured, queryable data access. & - Fine-tuning for embodied memory is computationally expensive. \newline - RAG requires efficient indexing and retrieval systems. \newline - Text-to-SQL generation may struggle with complex queries or dependencies. \\
\addlinespace
\textbf{Short-term Memory} & Acts as a temporary workspace within the LLM's context window, holding immediate contextual information for ongoing tasks. & - \textbf{Context Window Management}: Maintains recent conversational or input data within the transformer's limited context window. \newline - \textbf{Chunking and Summarization}: Breaks down large inputs into manageable pieces and condenses essential information to fit within the context window. & - Facilitates immediate contextual awareness. \newline - Chunking and summarization prevent information loss in long sequences. & - Limited by context window size, leading to truncation of older data. \newline - Summarization may omit critical details if not carefully designed. \\
\bottomrule
\end{tabular}
\label{tab:memory_components_part1}
\end{sidewaystable*}

\begin{sidewaystable*}[htbp]
\centering
\caption{Memory Components for LLM-Based Agents (Part 2)}
\begin{tabular}{>{\raggedright\arraybackslash}p{3cm} >{\raggedright\arraybackslash}p{3,2cm} >{\raggedright\arraybackslash}p{6cm} >{\raggedright\arraybackslash}p{3cm} >{\raggedright\arraybackslash}p{4cm}}
\toprule
\textbf{Component} & \textbf{Description} & \textbf{Key Techniques/\allowbreak Approaches} & \textbf{Advantages} & \textbf{Challenges/Limitations} \\
\midrule
\textbf{Data Storage Types} & Defines the types of information stored to support agent functionality. & - \textbf{Procedures (Agent Workflow Memory - AWM)}: Stores reusable task workflows derived from past experiences or queries to guide future actions. \newline - \textbf{Knowledge}: Includes external facts (e.g., articles, company rules) for context-specific responses. \newline - \textbf{User Information}: Stores personal user details (e.g., preferences, past activities) via systems like MemoryBank for personalized responses. & - Workflows improve efficiency by reusing successful routines. \newline - External knowledge enhances response accuracy. \newline - User information supports personalized interactions. & - Managing diverse data types requires robust storage systems. \newline - Privacy concerns with storing user information. \newline - Risk of outdated or irrelevant knowledge affecting performance. \\
\addlinespace
\textbf{Memory Management Issues} & Addresses challenges in storing and retrieving information efficiently. & - \textbf{Memory Duplication}: Consolidates similar records (e.g., combining successful action sequences into a unified plan or aggregating counts). & - Reduces redundancy and storage inefficiency. & - Duplication consolidation may lose nuanced details. \newline - FIFO overwriting risks losing valuable older data. \newline - Requires careful design to balance storage and retrieval efficiency. \\
\bottomrule
\end{tabular}
\label{tab:memory_components_part2}
\end{sidewaystable*}

\clearpage
With its robust memory system supporting processed observations and formulated plans, an LLM agent's operational flow then progresses to the execution system. This critical component is responsible for translating that internal understanding and knowledge into concrete interactions and actions within its environment.

\section{Execution System}
This system enables the agent to interact with its environment. It encompasses the mechanisms for tool orchestration, action invocation, and the immediate processing of action outcomes~\cite{xi2023risepotential}.
LLM agents interact with their environment and execute actions through several key mechanisms that bridge the gap between language understanding and real-world task automation~\cite{guo2024llm}. 
These mechanisms include:

\subsection{Tool and API Integration}
The most fundamental way LLM agents execute actions is through structured tool calling or function calling capabilities. Agents are given predefined functions, like file operations, database queries, web requests, or system commands, that correspond to particular actions they can perform. The agent generates structured outputs (typically JSON) that specify which tool to use and what parameters to provide. With this method, agents can carry out specific tasks like sending emails, generating files, performing computations, or getting data from other systems.~\cite{xi2023risepotential}.

\subsection{Multimodal Action Spaces}
Multimodal action spaces represent one of the most significant advances in LLM agent capabilities, enabling them to interact with environments beyond pure text interfaces~\cite{deng2023mind2web,zhou2024webarena}. 
Here's a deeper exploration:

\subsubsection{Visual Interface Automation:}
LLM agents can control graphical user interfaces through computer vision and automation frameworks to generate precise mouse clicks, keyboard inputs, and drag-and-drop operations~\cite{niu2024screenagent}. 
This capability allows agents to automate tasks in any software application, from web browsers to desktop applications, even when no programmatic API exists.
The technical implementation typically involves vision-language models that can process screenshots and generate coordinate-based actions, or integration with UI automation libraries that can identify elements through accessibility trees or DOM structures~\cite{rawles2023android}.

\subsubsection{Code Generation and Execution:}
A particularly powerful multimodal capability is dynamic code generation where agents write executable code in various programming languages to solve specific problems. This approach is especially valuable for data manipulation tasks, complex calculations, file processing, and integration between different systems. Agents can write Python scripts for data analysis, generate SQL queries for database operations, create shell scripts for system administration, or produce HTML/CSS/JavaScript for web-based solutions~\cite{gao2023pal,openai2023codeinterpreter}.

\subsubsection{Robotic and Physical System Control:}
In robotics applications, LLM agents can control physical systems through appropriate APIs and sensor integrations~\cite{xi2023risepotential}. They process sensor data (cameras, force sensors, temperature sensors) to understand the physical environment, generate motion plans and control commands, coordinate multiple actuators and subsystems, and adapt to real-time feedback from the physical world.

\subsection{Integration Challenges and Solutions}
Multimodal execution presents several technical challenges~\cite{guo2024llm}. 
Latency and coordination issues arise when combining different modalities, as visual processing and physical actions often require different timing considerations. Error propagation becomes more complex when failures can occur at multiple levels (perception, planning, execution). State synchronization requires careful management to ensure the agent's understanding remains consistent across different modalities~\cite{hwang2017seamlessintegrationcoordinationcognitive}.

\section{Discussion}
\subsection{Limitations}
While our review sheds light on the foundational elements of intelligent LLM agents, several limitations warrant consideration. Firstly, these agents currently fail at certain operations that humans can easily perform, largely due to a lack of sufficient experience interacting in specific environments. Teaching these experiences to LLMs is exceptionally costly, often requiring extensive fine-tuning. This challenge is compounded by the fact that many advanced models are closed-source, making it difficult to fine-tune this models. Moreover, acquiring the necessary data for targeted training is also time-consuming. Secondly, while LLMs excel at generating and understanding text, their ability to generate precise actions in the real world or within graphical user interfaces (GUIs) remains limited. Thirdly, despite advancements, visual perception in these agents is not yet as robust as required, with many mistakes stemming from an incomplete or inaccurate understanding of the environment.

\subsection{Implications}
The review presented in this paper has significant implications for the future of artificial intelligence. By demonstrating that LLM agents can move beyond simple language generation to exhibit capabilities akin to human cognition, we open doors for their application in highly complex domains requiring nuanced understanding and decision-making, such as scientific discovery, personalized education, and advanced robotics. The modular design and the integration of specialized components suggest a promising path towards building more robust and adaptable AI systems that can learn and evolve. Furthermore, the memory capabilities highlighted in this review could lead to the development of AI assistants that are not only more helpful but also more reliable and context-aware.

\subsection{Possible Extensions}
Future research can extend this work in several promising directions. One critical area is to explore more advanced mechanisms for knowledge acquisition and self-correction in LLM agents, enabling them to continuously learn from new experiences and rectify errors without extensive human intervention. However, it would also be very interesting to investigate how these agents can learn to accomplish a task after just a single demonstration with human help, subsequently performing it autonomously. 
This~\enquote{learn-from-one-shot} paradigm could significantly reduce the cost and effort of training LLM agents in new domains. An even more ambitious extension could be developing agents where humans act as assistants. This would improve productivity by 10x.

\section{Conclusion}
This paper set out to explore the intricate design and implementation strategies for creating intelligent LLM agents, focusing on their core capabilities across perception, memory, reasoning, planning, and execution. Our exploration revealed that LLM agents are not merely large language models, but complex systems built upon specialized components that mimic human cognitive processes. Specifically, we reviewed reasoning techniques, such as Chain-of-Thought and Tree-of-Thought, that significantly enhance an agent's problem-solving abilities. 

Moreover, the review showed that using different experts to focus on each part of the reasoning improves performance. Another conclusion from the review is that robust memory systems are crucial for personalized responses, continuous learning, and long-term coherence and adaptability. 

Furthermore, our analysis highlighted the critical role of a well-implemented perception system in enabling agents to interpret diverse environmental inputs, and the necessity of action systems for translating decisions into tangible outcomes. These findings directly address our initial objectives by illustrating how specific architectural designs and advanced techniques contribute to building more capable and generalized LLM agents, moving beyond simple workflow automation towards truly autonomous and intelligent entities.

\newpage
\bibliographystyle{splncs04}


\end{document}